\documentclass[lettersize,journal]{IEEEtran}
\usepackage{amsmath,amsfonts}
\usepackage{algorithmic}
\usepackage{algorithm}
\usepackage{array}
\usepackage{textcomp}
\usepackage{stfloats}
\usepackage{url}
\usepackage{verbatim}
\usepackage{graphicx}
\usepackage{cite}
\usepackage{booktabs}
\usepackage{multirow}
\usepackage{xcolor}
\usepackage{adjustbox}   
\usepackage{tabularx}
\usepackage{threeparttable}
\usepackage{array}
\usepackage{rotating}
\usepackage{bm}
\usepackage{subfigure}

\usepackage{xcolor}  
\usepackage{hyperref}  
\hypersetup{
    colorlinks=true,    
    linkcolor=blue,     
    citecolor=blue,     
    urlcolor=blue       
}

\begin{document}

\title{

When Prompting Meets Spiking: Graph Sparse Prompting via Spiking Graph  Prompt Learning}

\author{Bo Jiang, Weijun Zhao, Beibei Wang, Jin Tang 
\thanks{The authors are with the School
of Computer Science and Technology, Anhui University, Hefei 230601, China}
\thanks{Manuscript received April 19, 2021; revised August 16, 2021.}}

\markboth{Journal of \LaTeX\ Class Files,~Vol.~14, No.~8, August~2021}%
{Shell \MakeLowercase{\textit{et al.}}: A Sample Article Using IEEEtran.cls for IEEE Journals}


\maketitle

\begin{abstract}

Graph Prompt Feature (GPF) learning has been widely used in adapting pre-trained GNN model on the
downstream task. 
GPFs first introduce some prompt atoms and then learns the optimal prompt vector for each graph node using the linear combination of prompt atoms. 
However, existing GPFs generally conduct prompting over  node's all feature dimensions which is obviously redundant and also be sensitive to node feature noise. 
To overcome this issue, for the first time, this paper proposes learning \emph{sparse} graph prompts by leveraging the \emph{spiking neuron} mechanism, termed  Spiking Graph Prompt Feature (SpikingGPF). Our approach is motivated by the observation that spiking neuron can perform inexpensive information processing and produce sparse outputs which naturally fits the task of our graph sparse prompting. 
Specifically,
SpikingGPF has two main aspects. 
First, it learns a sparse prompt vector for each node by exploiting a spiking neuron architecture, enabling prompting on selective node features. This yields a more compact and lightweight prompting design while also improving robustness against node noise. 
Second, SpikingGPF introduces a novel prompt representation learning model based on sparse representation theory, i.e., it  represents each node prompt as a sparse combination of prompt atoms. This encourages a more compact representation and also facilitates efficient computation.
Extensive experiments on several benchmarks demonstrate the effectiveness and robustness of SpikingGPF. 

\end{abstract}


\section{Introduction}
\label{Introduction}
\IEEEPARstart{G}{raph} Neural Networks 
(GNNs)~\cite{gilmer2017neural,kipf2016semi} have transformed relational data analysis through message-passing mechanisms that capture complex dependencies in graph structures. 
This capability has enabled widespread applications in social network analysis~\cite{wang2018billion}, recommendation systems~\cite{zhou2021temporal}, and Web mining~\cite{agarwal2022graphnli}. 
Existing GNNs often rely on supervised learning, which requires large amounts of task-specific labeled data~\cite{ma2024hetgpt}. Obtaining such labeled data is usually time-consuming and expensive, thereby limiting the practical application of GNNs in many real-world scenarios. 
To overcome this issue, 
the ‘pre-training, fine-tuning’ paradigm has attracted increasing attention in recent years. 
This paradigm first learns general, task-agnostic knowledge from a large amount of unlabeled graphs, and then fine-tunes the pre-trained models on various downstream tasks using task-specific labels~\cite{yu2025non}. However, the semantic misalignment between pre-training and downstream graphs can lead to suboptimal adaptation or negative transfer, degrading generalization~\cite{huang2024measuring}. Moreover, full-parameter fine-tuning of large models is computationally expensive in terms of memory and time.

Recently, inspired by prompt tuning in natural language processing (NLP)~\cite{liu2023pre} and computer vision (CV)~\cite{jia2022visual}, researchers have  explored the prompt based fine-tuning in the graph learning field. 
Graph prompt learning aims to fine-tune a pre-trained model on the downstream task while keeping its parameters frozen by introducing only a small set of learnable prompt vectors~\cite{jiang2024reliable}. 
Overall, existing graph prompt learning methods can be broadly categorized into two types: task-level prompting and data-level prompting.
For task-level prompting, some works focus on designing task templates that align pre-training objective and downstream task. For example, GPPT~\cite{sun2022gppt} transforms the downstream node classification problem into the pre-trained link prediction task. 
GraphPrompt~\cite{liu2023graphprompt} unifies pre-trained link prediction with downstream node- and graph-classification tasks into a subgraph similarity task. 
For data-level prompting, some studies aim to adapt downstream task by modifying the input graph node or edge representation.
For example, GPF~\cite{fang2023universal} augments node features with learnable prompt vectors, while EdgePrompt~\cite{Fu2025EdgePT} learns tunable prompts for the edges of the input graph. SUPT~\cite{SUPT} introduces prompt features at the subgraph level to guide the representations of pre-trained GNNs. VNT~\cite{VNT} injects a set of trainable virtual nodes as soft prompts into the pre-trained graph Transformer encoder.

However, a critical limitation lies in the above existing data-level prompting methods: they universally employ a prompting strategy across all graph elements (e.g., nodes, edges, attributes). This strategy is obviously redundant and often sub-optimal~\cite{zhu2025relief,jiang2024reliable}.  
For prompt learning problem, it is 
obviously desirable to obtain more \emph{lightweight} and \emph{compact} prompt representation to  `augment/modify'  the input graph data. 
Based on this consideration, 
we incorporate the \textbf{sparse constraints} into graph prompt learning and develop sparse prompting for graph fine-tuning. 
As we all know that spiking neuron learning mechanism can
inherently obtain sparse outputs with inexpensive cost~\cite{zhu2022spiking,WangJTBL25}. 
Inspired by this, we think
that spiking neuron mechanism naturally fits the task of
our graph sparse prompting. 

Specifically, in this paper, we mainly focus on graph node prompting problem and propose a novel Spiking Graph Prompt Feature method (SpikingGPF) for graph fine-tuning. Our SpikingGPF follows the pipeline of the widely used Graph Prompt Feature (GPF)~\cite{fang2023universal} which first introduces some prompt atoms (basis prompt vectors) $\textbf{B}=\{\textbf{b}_1,\textbf{b}_2\cdots \textbf{b}_K\}$ and
then learns the prompt $\textbf{p}_i$ for each node $v_i$ by using the linear combination
of these prompt atoms. 
\emph{In contrast}, the core aspect of our SpikingGPF is to learn an optimal \textbf{sparse prompt} $\textbf{p}_i$ by leveraging the spiking neuron mechanism, which can select some optimal feature dimensions for prompting. 
Moreover, 
in SpikingGPF, each node prompt $\textbf{p}_i$ is represented not  by the full atom set $\textbf{B}$ (as conducted in GPF~\cite{fang2023universal}), but by a linear combination of a \textbf{sparsely selected subset} of  $\textbf{B}$. This allows each node to focus on its most relevant atoms and also enables more efficient computation. 
On the whole, when compared with GPF~\cite{fang2023universal}, our SpikingGPF is obviously more lightweight and compact while also performs more robustly with respect to graph node noise. 

The main contributions of this paper are summarized as follows:

\begin{itemize}
\item 
We propose a novel spiking prompt learning mechanism, termed Spiking Graph Prompt Feature (SpikingGPF) by developing a new spiking neuron architecture to learn graph prompt sparsely and robustly. 
To our best knowledge, this work is the first attempt to leverage spiking model for graph sparse prompt learning problem. 

\item 
We propose a new prompt representation  method 
by leveraging sparse representation theory, i.e., each node prompt is {sparsely} represented by linear combination of prompt atoms, encouraging more compact and efficient learning of node prompts. 

\item
Experimental results on various benchmark datasets
demonstrate the effectiveness and benefits of the proposed SpikingGPF on various pre-trained GNN models.  
\end{itemize}




\section{Related Work}

\subsection{Graph pre-training}
Graph pretraining leverages unlabeled graph data in a self-supervised manner to encode structural and semantic information, thereby improving performance on downstream tasks. It is usually divided into two types, i.e., contrastive  methods and generative  methods. Contrastive methods encourage consistency between different views (or substructures) of the same graph, thereby learning discriminative embeddings. For example, GraphCL~\cite{you2020graph} maximizes the agreement between different augmented views of the same graph. SimGRACE~\cite{xia2022simgrace} generates views by perturbing model parameters, enabling contrastive learning without graph augmentations. DGI~\cite{velivckovic2018deep} and InfoGraph~\cite{sun2019infograph} learn node-level embeddings by maximizing the mutual information between global graph summary and local subgraph representations.
On the other hand, generative methods pretrain models to recover masked or corrupted parts of the graph, thereby forcing encoders to capture useful structural and semantic information. For instance, GraphMAE pre-trains GNNs by reconstructing masked node features. EdgePred~\cite{liu2023graphprompt,sun2022gppt} captures structural patterns in local neighborhoods by predicting missing edges in the graph.

\subsection{Prompt-based learning}
Recently, graph prompt learning~\cite{fu2025graph} has gained popularity as an alternative to fine-tune pre-trained GNN models.
Unlike traditional fine-tuning pipeline, it keeps the pre-trained model parameters unchanged and uses a small number of learnable prompt vectors to guide the adaptation on the downstream tasks.
Broadly speaking, existing methods fall into two categories: task-level prompting and data-level prompting. 

\textbf{Task-level prompting.}
Task-level prompting reformulates downstream tasks to align their objective space with that of pre-training tasks, thereby mitigating performance degradation caused by intrinsic domain discrepancies.  For instance, GPPT~\cite{sun2022gppt} uses link prediction as pre-training task, and re-formulates the downstream node classification task into a pre-training task by converting nodes into token pairs. All-in-One~\cite{sun2023all} constructs induced graphs over nodes and edges, thereby transforming node-level and edge-level tasks into graph-level tasks. GraphPrompt~\cite{liu2023graphprompt} treats subgraph similarity as a universal task template to unify pre-training and downstream tasks, and also introduces task-specific prompts to assist downstream adaptation. HetGPT~\cite{ma2024hetgpt} designs virtual class prompt and heterogeneous feature prompt to directly align pre-training and downstream task objective.
OFA~\cite{OFA} introduces the concepts of Nodes-of-Interest (NOI) subgraphs and prompt nodes, transforming  various downstream tasks into a unified binary classification task. 

\textbf{Data-level prompting.}
Inspired by prevalent prompting methods in natural language processing~\cite{liu2023pre}, an intuitive graph prompting strategy operates at the graph data level. The intuition of data-level prompting is to modify the input graph data to fit specific downstream tasks. For example, GPF and GPF-plus~\cite{fang2023universal} propose a universal prompt-tuning method that injects additional learnable prompt vectors into the node features on the downstream tasks, thereby converting the input graph into a prompted graph. EdgePrompt~\cite{Fu2025EdgePT} designs graph prompts from an edge perspective, introducing learnable prompt vectors on edges during message passing. 
VNT~\cite{VNT} employs trainable virtual nodes as soft prompts on graphs, enabling fast adaptation to downstream tasks with only a few labeled samples, while keeping the pre-trained encoder frozen.
RELIEF~\cite{zhu2025relief} formulates “which nodes to prompt” and “what prompts to add” as a combinatorial optimization problem, and using reinforcement learning to automatically search for the optimal prompting strategy to generate the prompted graph.

\section{Preliminaries}
 
Given the graph data $\mathcal{G}=(\textbf{X},\textbf{A})$ with $\textbf{X}\in\mathbb{R}^{n\times d}$ denoting the node features 
and $\textbf{A}\in\mathbb{R}^{n\times n}$ denoting the adjacency matrix,  based on pre-trained  model $f_{\Phi}$, 
the general graph prompt learning problem on the downstream task can be formulated as~\cite{fang2023universal,jiang2024reliable}, 
\begin{equation}
    \label{eq:opt}
    \min_{\theta,\;\mathcal{P}}\; \mathcal{L}^{down} \,\bigl(f_{\Phi}(\tilde{\mathcal{G}}({\textbf{X}},{\textbf{A}};\mathcal{P})),\;\theta\bigr),
\end{equation}
where $\tilde{\mathcal{G}}$ denotes the prompted graph and $\mathcal{P}$ denotes the learnable prompt parameters. $\theta$ denotes the trainable classifier head for the downstream task. 

Graph Prompt Feature (GPF)~\cite{fang2023universal} is a commonly used graph fine-tuning method by conducting prompt learning in graph node attribute/feature space.
It  introduces several learnable prompt atoms (basis vectors) $\textbf{B}=\{\textbf{b}_1,\textbf{b}_2\dots\textbf{b}_K\}\in\mathbb{R}^{K\times d}$ and learns the node prompt $\textbf{p}_i$ by using the linear combination of these prompt atoms.  
Specifically, 
it first computes the coefficient $\textbf{s}_{ik}$ as 
\begin{align}
    \textbf{s}_{ik} =
    \frac{\exp(\textbf{w}_k \textbf{x}^T_i)}{\sum_{l=1}^{K} \exp(\textbf{w}_l \textbf{x}^T_i)}
\end{align}
where $\textbf{W}=\{ \textbf{w}_1,\textbf{w}_2\cdots\textbf{w}_K \}\in\mathbb{R}^{K\times d}$ are learnable parameters.  
Then, GPF~\cite{fang2023universal} obtains the prompt  $\textbf{p}_i$ for each node $v_i$ by using the linear combination of all prompt atoms in $\textbf{B}$ as
\begin{equation}
    \textbf{p}_i = \sum^K_{k=1}\textbf{s}_{ik}\textbf{b}_k
    \label{eq:pi_define}
\end{equation}
%
Finally, the prompt atoms $\textbf{B}$ as well as parameters $\{\textbf{W},\theta\}$ 
are learned by optimizing the objective on the downstream task as, 
\begin{align}\label{eq:finally_opt}
&\min_{\textbf{B},\,\textbf{W},\theta}\,\,\mathcal{L}^{down}\,\big(f_{\Phi}\big(\tilde{\mathcal{G}}({\textbf{X}}+\textbf{P},\textbf{A}; \textbf{B},\textbf{W})\big), \;\theta \big) 
\nonumber
\end{align}
where $\textbf{P}=\{\textbf{p}_1\cdots \textbf{p}_n\}$ and $\textbf{p}_i$ is defined via Eq.(\ref{eq:pi_define}).

\section{Graph Spiking Prompt Feature}
\label{Method}
The above GPF generally
suffers from two main issues. 
First, it obtains the prompt $\textbf{p}_i$ as the linear combination of \emph{all} prompt atoms which are redundant and also induces high computation complexity. 
Moreover, it adopts the deterministic prompting mechanism on each node's all attributes (features). This ‘complete’ prompting mechanism is obviously redundant
and also sensitive w.r.t node feature noise. 

To address these issues, 
we propose a new  more lightweight and compact
prompting (termed SpikingGPF) by incorporating \emph{sparse constraints} into the above GPF model~\cite{fang2023universal}.  
Specifically, in SpikingGPF, we first represent each prompt $\textbf{p}_i$ by selecting a subset of atoms from $\textbf{B}$ and using linear combination of these selected atoms. 
This can be achieved by enforcing the coefficients $\textbf{S}$ to be sparse. 
Then, we encourage each prompt $\textbf{p}_i$ to be sparse, 
i.e., for each node, we select some optimal feature dimensions for
prompting, allowing the model to focus on some most relevant features. 
Below, we introduce these two modules respectively.


\subsection{Learning sparse coefficient $\textbf{S}$}

\begin{figure}[!hptb]
    \centering
    \includegraphics[width=0.95\linewidth]{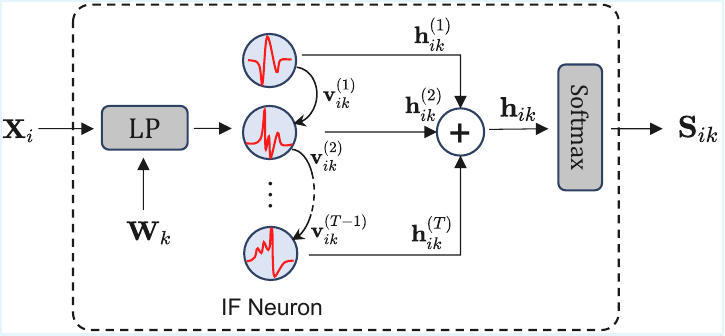}
    \caption{Illustration of the learning architecture of our $\textbf{S}$-learning module.
    }
    \label{fig:S_learning_framework}
\end{figure}

As shown in Fig.~\ref{fig:S_learning_framework}, our $\textbf{S}$-learning 
module includes Linear Projection (LP) layer and  Integrate-and-Fire (IF) neuron layer. 
Specifically, 
taking node feature $\textbf{x}_i$ as input, the LP layer outputs $\bm \alpha_{ik}$ using learnable parameter weight vector $\textbf{w}_k$ as,
\begin{equation}
\bm \alpha_{ik}=\textbf{w}_k\textbf{x}^T_i
\end{equation}
Then, $\bm \alpha_{ik}$
charges several IF neurons~\cite{zhu2022spiking,WangJTBL25} each of which conducts integrate, fire and reset operation respectively.



\begin{itemize}
    \item \textit{Integrate.}
    Given $\textbf{v}^{(0)}_{ik}=0$, integration operation is to 
    update the membrane potential, i.e.,  
         \begin{equation}
         \label{eq:SSA_integrate}
            \tilde{\textbf{v}}^{(t)}_{ik} =\Psi(\textbf{v}^{(t-1)}_{ik}, \bm \alpha_{ik})= \textbf{v}^{(t-1)}_{ik} + \bm \alpha_{ik}
        \end{equation}
        where 
        $t=1,2\cdots T$ denotes the $t$-th IF neuron.
    \item \textit{Fire.}
    Fire action is used when the accumulated membrane potential exceeds a spiking threshold  $\mu$, i.e.,  
        \begin{equation}
        \label{eq:SSA_fire}
            \textbf{h}^{(t)}_{ik} = \mathcal{H}( \tilde{\textbf{v}}^{(t)}_{ik} - \mu) =
            \begin{cases} 
                \begin{alignedat}{2}
                    & 1 \qquad  &&\text{if} \quad \tilde{\textbf{v}}^{(t)}_{ik} \ge \mu,        \\
                    & 0 \qquad  &&\text{otherwise}. 
                \end{alignedat}
            \end{cases}
        \end{equation}
    \item \textit{Reset.}
        After firing, IF neuron needs to reset the membrane potential. We use the soft reset function~\cite{zhu2022spiking,li2023graph} which is defined as, 
        \begin{equation}
        \label{eq:SSA_reset}
            \textbf{v}^{(t)}_{ik} = 
            \mathcal{R}(\tilde{\textbf{v}}^{(t)}_{ik}, \textbf{h}^{(t)}_{ik}, \mu) =
            \tilde{\textbf{v}}^{(t)}_{ik} - \mu \;  \textbf{h}^{(t)}_{ik}
        \end{equation}
        where ${\textbf{v}}^{(t)}_{ik}$ is used for the next  IF neuron.
\end{itemize}
%

Finally, we perform average aggregation~\cite{zhu2022spiking} of the outputs from all $T$ spiking neurons to obtain the final output as
\begin{equation}
    \textbf{h}_{ik} = \frac1{T}\sum_{t=1}^T \textbf{h}^{(t)}_{ik} 
\label{eq:h_avg}
\end{equation}
The coefficients $\textbf{S}$ are obtained by further conducting the Softmax operation on $\textbf{H}$ as, 
\begin{equation}
    \label{eq:s_define}
    \textbf{s}_{ik} = \frac{\text{exp}(\textbf{h}_{ik})}{\sum_{j=1}^K \text{exp}(\textbf{h}_{ij})}.
\end{equation}

\begin{figure}[!ht]
    \centering
    \subfigure[\scriptsize Visualizations of learned $\textbf{S}$ under different $\mu$ values ($T=4$).]{
        \includegraphics[width=0.46\linewidth]{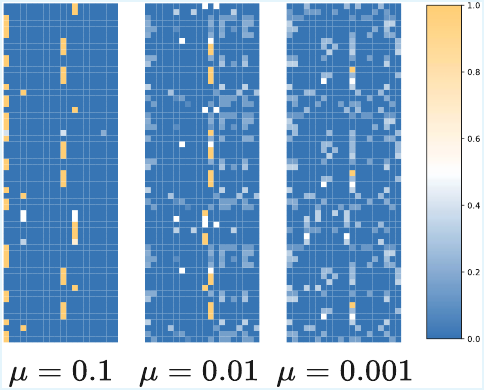}
    }
    \hfill
    \subfigure[\scriptsize Visualizations of learned $\textbf{S}$ under different $T$ values ($\mu=0.01$).]{
        \includegraphics[width=0.46\linewidth]{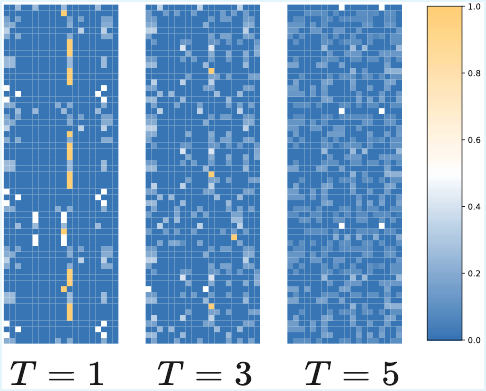}
    }
    \caption{Visualizations of learned $\textbf{S}$ ($n=59, K=20$ in this example) on KarateClub~\cite{rozemberczki2020karate} dataset.}
    \label{fig:vis_S}
\end{figure}

 \textbf{Remark.}
As shown in Eq.(\ref{eq:SSA_fire}), the output $\textbf{h}^{(t)}_{ik}$ of each  IF neuron
is sparse whose sparsity is  controlled via the threshold $\mu$.
As shown in Eqs.(\ref{eq:h_avg},\ref{eq:s_define}), since $\textbf{S}$ is the average combination of $\textbf{h}^{(t)}_{ik}$ as well as softmax operation, it is also encouraged to be sparse when $T$ is not very large. Therefore, the sparsity of the learned $\textbf{S}$ is determined by the parameters $\{\mu,T\}$. 
Figure~\ref{fig:vis_S} illustrates some visualizations  of learned $\textbf{S}$ under different parameter $\mu$ and $T$ values, respectively. We can note that higher threshold $\mu$ and smaller $T$ can encourage more sparse $\textbf{S}$ solution. This is consistent with the above analysis which clearly demonstrates the desired sparsity of learned  $\textbf{S}$. 

\subsection{Learning sparse prompt $\textbf{p}_i$}


After we obtain the sparse $\textbf{S}$, as conducted in Eq.(\ref{eq:pi_define}), we 
can generate prompt $\textbf{p}_i$ for node $v_i$ via the linear combination of prompt atoms as
\begin{equation}
\textbf{p}_i=\sum^{K}_{k=1}\textbf{s}_{ik}\textbf{b}_k
\label{eq:gpf_combination}
\end{equation}
Since $\textbf{S}$ is sparse, 
the optimal \emph{subset} of $\textbf{B}=\{\textbf{b}_1,\textbf{b}_2\cdots \textbf{b}_K\}$ is selected for learning $\textbf{p}_i$. 
However, 
as discussed in Introduction, our goal here is to further 
learn sparse prompt $\textbf{p}_i$, i.e., encouraging some elements of $\textbf{p}_i$ to be zero. 
Obviously, the above linear combination Eq.(\ref{eq:gpf_combination}) cannot encourage the sparse constraint of learned $\textbf{p}_i$. 

\begin{figure}[h]
    \centering
    \includegraphics[width=0.75\linewidth]{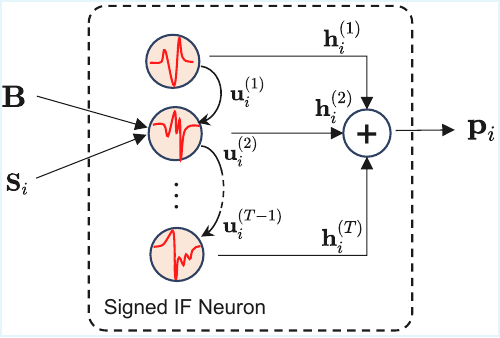}
    \caption{Illustration of the learning architecture of our \textbf{P}-learning module.
    }
    \label{fig:P_learning_framework}
\end{figure}

To achieve this purpose,  
as shown in Fig.~\ref{fig:P_learning_framework}, we propose our $\textbf{P}$-learning module by adopting 
a signed IF mechanism~\cite{kim2020spiking} which 
takes the prompt atom set $\textbf{B}$ and $\textbf{s}_i$ (the $i$-th of $\textbf{S}$) as input and 
employ 
several signed IF neurons to obtain sparse $\textbf{p}_i$. 
Each signed IF neuron involves 
integrate, signed fire and reset operation respectively. 
\begin{itemize}
    \item \textit{Integrate.} 
It is used to update the membrane potential vector $\textbf{u}^{(t-1)}_{i}$ as, 
    \begin{equation}
        \tilde{\textbf{u}}^{(t)}_{i} = \Psi(\textbf{u}^{(t-1)}_{i}, \textbf{s}_{i}) = \textbf{u}^{(t-1)}_{i} + \textbf{s}_{i}\textbf{B}
    \end{equation}
where 
$t=1,2\cdots T$ denotes the $t$-th signed IF neuron.  
    \item \textit{Signed Fire.} 
    Inspired by~\cite{kim2020spiking}, we adopt a signed fire action which is defined as
    \begin{equation}
        {\textbf{h}}^{(t)}_{id} = 
        \tilde{\mathcal{H}}( \tilde{\textbf{u}}^{(t)}_{id} - \gamma ) =
                \begin{cases} 
                    \begin{alignedat}{2}
                        & +1 \qquad  &&\text{if} \quad \tilde{\textbf{u}}^{(t)}_{id} \ge \gamma        \\
                        & -1 \qquad  &&\text{if} \quad \tilde{\textbf{u}}^{(t)}_{id} \le - \gamma        \\
                        & 0 \qquad  &&\text{otherwise}. 
                    \end{alignedat}
                \end{cases} \\
    \label{eq:h_id^t}
    \end{equation}
    where $\gamma$ denotes the firing threshold.
    
    \item \textit{Reset.} After firing operation, it needs to reset the membrane potential vector as, 
        \begin{equation}
        \label{eq:SSP_reset}
            \textbf{u}^{(t)}_{i} =
            \mathcal{R}(\tilde{{\textbf{u}}}^{(t)}_{i}, \textbf{h}^{(t)}_{i}, \gamma)=
            \tilde{{\textbf{u}}}^{(t)}_{i} - \gamma \; \textbf{h}^{(t)}_{i} 
        \end{equation}
\end{itemize}
Finally, 
we take the average of the $T$ neuron outputs to obtain the final prompt vector $\textbf{p}_i$ as, 
\begin{equation}
    \textbf{p}_{i} = \frac1{T}\sum_{t=1}^T {\textbf{h}}^{(t)}_{i} 
    \label{eq:pi_avg}
\end{equation}

\begin{figure}[!ht]
    \centering
    \subfigure[Visualizations of $\textbf{P}$ under different $\gamma$ values  ($T = 3$)]{
        \includegraphics[width=0.95\linewidth]{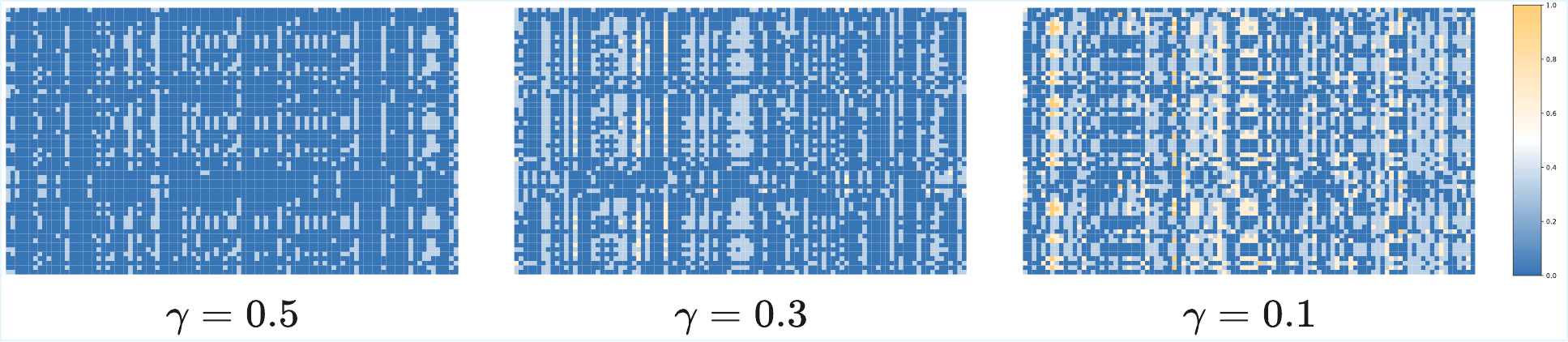}
    }
    \hfill
    \subfigure[Visualizations of $\textbf{P}$ under different $T$ values ($\gamma = 0.35$)]{
        \includegraphics[width=0.95\linewidth]{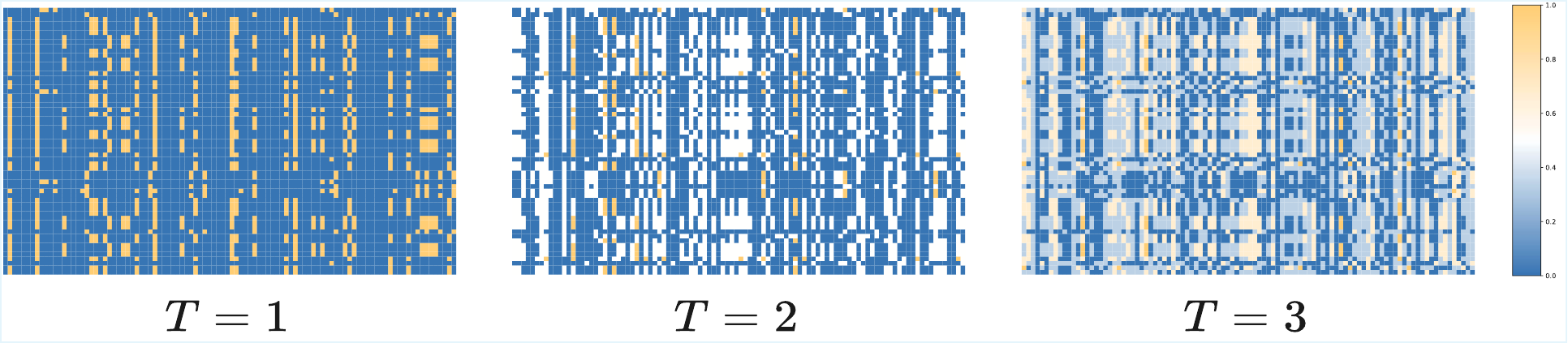}
    }
   \caption{Visualizations of learned prompts $\textbf{P}$ on KarateClub~\cite{rozemberczki2020karate} dataset under different parameter $\gamma$ and $T$ values.}
    \label{fig:vis_P}
\end{figure}

\textbf{Remark.}
As shown in Eq.(\ref{eq:h_id^t}), the output $\textbf{h}^{(t)}_{id}$ of each signed IF neuron 
is sparse whose sparsity is determined via threshold $\gamma$. Thus, it encourages the sparsity of the 
output $\textbf{p}_{i}$ (Eq.(\ref{eq:pi_avg}))  when $T$ is not very large, i.e., the sparsity of the learned
$\textbf{p}_i$ is determined via parameters $\gamma$ and $T$. 
Fig.~\ref{fig:vis_P} illustrates the example of learned $\textbf{p}_i$ across different
$\gamma$ and $T$ respectively. This further demonstrates the desired sparsity
of $\textbf{p}_i$ based on the above learning mechanism.

After the sparse prompts $\mathrm{\textbf{P}} = \{\textbf{p}_1, \textbf{p}_2 \cdots \textbf{p}_{n} \}$ are learned, we obtain the prompted graph by adding them to the original node features $\textbf{X}$. 
The prompted graph is then fed into the frozen pre-trained encoder 
$f_\Phi$ for the downstream task prediction. 
The prompt atoms $\textbf{B}$ as well as parameters $\{\textbf{W},\theta\}$ 
are learned by optimizing the following objective on the downstream task,  
\begin{align}\label{eq:finally_opt}
&\min_{\textbf{B},\,\textbf{W},\theta}\,\,\mathcal{L}^{down}\,\big(f_{\Phi}\big(\tilde{\mathcal{G}}({\textbf{X}}+\textbf{P},\textbf{A}; \textbf{B},\textbf{W})\big), \;\theta \big)  \\
& s.t. \,\,{\textbf{P}}=\textbf{S}\textbf{B}^T.
\nonumber
\end{align}

\section{Experiments}

In this section, we conduct experiments to evaluate the advantages of SpikingGPF on various benchmark datasets.
\begin{table}[!htbp]
    \small
    \centering
    \caption{Summary of All Datasets Used in Our Experiments.}
    \label{table:dataset}
    \renewcommand{\arraystretch}{1.2} 
    \begin{tabular}{l|cccccc}
    
    \toprule
    Dataset           & Nodes  & Edges  & Features  & Classes   \\ 
    
    \midrule
    
    Cora                     & 2,708      & 5,429      & 1,433      & 7        \\ 
    CiteSeer                 & 3,327      & 4,732      & 3,703      & 6        \\
    Pubmed                   & 19,717     & 44,338     & 500        & 3        \\
    Photo                    & 7,487      & 119,043    & 745        & 8        \\
    Computers                & 13,381     & 245,778    & 767        & 10       \\
    CS                       & 18,333     & 81,894     & 6,805      & 15       \\
    Physics                  & 8,415      & 34,493     & 247,962    & 5        \\
    Wisconsin                & 251        & 515        & 1,703      & 5        \\
    Texas                    & 183        & 325        & 1,703      & 5        \\
    Cornell                  & 183        & 298        & 1,703      & 5        \\
    Ogbn-arxiv               & 169,343    & 1,166,243  & 128        & 40      \\ 
       
    \bottomrule
        
    \end{tabular}
\end{table}

\subsection{Experimental Setup}
\textbf{Datasets.} 
To evaluate our proposed SpikingGPF, we test it on $11$ widely used benchmark datasets, including four citation networks~\cite{sen2008collective,hu2020open}, 
two Amazon co-purchase graphs~\cite{shchur2018pitfalls}, 
two coauthor networks~\cite{shchur2018pitfalls}, 
three heterophilic web page datasets~\cite{pei2020geom}.  
The details of these datasets are listed in Table~\ref{table:dataset}.

\begin{table*}[!htpb]
\small
\centering
\caption{Accuracy of 1-shot node classification.}
\label{tab:node_1shot}
\begin{adjustbox}{width=\textwidth}
\renewcommand{\arraystretch}{1.6} 
\begin{tabular}{@{}l | l | *{11}{>{\raggedright\arraybackslash}p{1.5cm}} @{}}
\toprule
\multicolumn{2}{c}{\textbf{Methods $\backslash$ Datasets}}
  & \multicolumn{1}{c}{\textbf{Cora}}
  & \multicolumn{1}{c}{\textbf{CiteSeer}}
  & \multicolumn{1}{c}{\textbf{PubMed}}
  & \multicolumn{1}{c}{\textbf{Photo}}
  & \multicolumn{1}{c}{\textbf{Computers}}
  & \multicolumn{1}{c}{\textbf{CS}}
  & \multicolumn{1}{c}{\textbf{Physics}}
  & \multicolumn{1}{c}{\textbf{Wisconsin}}
  & \multicolumn{1}{c}{\textbf{Texas}}
  & \multicolumn{1}{c}{\textbf{Cornell}}
  & \multicolumn{1}{c}{\textbf{Ogbn-arxiv}} \\
\midrule

\multicolumn{2}{c|}{GCN} & $45.92_{\pm6.14}$ & $32.17_{\pm12.59}$ & $57.14_{\pm4.21}$ & $40.18_{\pm11.40}$ & $27.07_{\pm6.64}$ & $47.94_{\pm4.89}$ & $59.20_{\pm16.14}$ & $20.63_{\pm7.88}$ & $26.46_{\pm19.07}$ & $21.37_{\pm6.24}$ & $7.64_{\pm1.96}$ \\
\midrule
\multicolumn{2}{c|}{GAT} & $45.73_{\pm3.52}$ & $32.31_{\pm12.74}$ & $57.60_{\pm7.53}$ & $24.49_{\pm13.11}$ & $13.98_{\pm5.53}$ & $32.65_{\pm5.11}$ & $51.27_{\pm18.42}$ & $17.30_{\pm6.90}$ & $27.33_{\pm14.79}$ & $21.24_{\pm4.67}$ & $8.89_{\pm1.27}$ \\
\midrule
\multirow{6}{*}{\rotatebox{90}{\textbf{GraphCL}}}
  & Fine-Tuning & $44.04_{\pm5.81}$ & $30.53_{\pm10.63}$ & $56.42_{\pm3.84}$ & $40.69_{\pm12.61}$ & $25.22_{\pm6.71}$ & $51.23_{\pm5.05}$ & $65.34_{\pm11.02}$ & $24.23_{\pm8.15}$ & $25.09_{\pm16.61}$ & $21.24_{\pm4.50}$ & $4.42_{\pm1.00}$ \\
  & All-in-One & $41.02_{\pm5.55}$ & $31.44_{\pm7.40}$ & $51.77_{\pm6.17}$ & $33.80_{\pm12.31}$ & $29.46_{\pm11.11}$ & $35.81_{\pm9.32}$ & $36.17_{\pm9.65}$ & $23.06_{\pm6.44}$ & $25.22_{\pm6.04}$ & $21.74_{\pm7.77}$ & $3.39_{\pm2.53}$ \\
  & GPF & $47.91_{\pm6.72}$ & $34.77_{\pm13.46}$ & $60.45_{\pm5.49}$ & $43.41_{\pm11.53}$ & $26.11_{\pm7.34}$ & $57.78_{\pm8.19}$ & $73.91_{\pm7.33}$ & $20.36_{\pm8.53}$ & $27.33_{\pm16.54}$ & $23.73_{\pm6.68}$ & $4.87_{\pm0.98}$ \\
  & GPF-plus & $46.43_{\pm7.19}$ & $33.73_{\pm11.74}$ & $57.76_{\pm4.10}$ & $41.97_{\pm7.50}$ & $27.00_{\pm6.83}$ & $55.64_{\pm6.81}$ & $71.99_{\pm9.93}$ & $18.74_{\pm7.02}$ & $26.34_{\pm17.75}$ & $24.97_{\pm8.60}$ & $8.95_{\pm2.52}$ \\
  & SpikingGPF & $50.30_{\pm5.01}$ & $40.76_{\pm8.87}$ & $61.52_{\pm6.67}$ & $45.70_{\pm7.37}$ & $32.26_{\pm9.45}$ & $62.48_{\pm3.72}$ & $75.26_{\pm4.75}$ & $25.59_{\pm6.42}$ & $31.80_{\pm15.83}$ & $24.60_{\pm7.77}$ & $11.18_{\pm2.03}$ \\
\midrule
\multirow{6}{*}{\rotatebox{90}{\textbf{SimGRACE}}}
  & Fine-Tuning & $35.17_{\pm4.79}$ & $28.65_{\pm12.24}$ & $55.24_{\pm4.92}$ & $32.88_{\pm11.12}$ & $24.31_{\pm6.88}$ & $43.68_{\pm5.78}$ & $65.04_{\pm10.69}$ & $17.66_{\pm8.60}$ & $27.33_{\pm14.51}$ & $23.48_{\pm5.29}$ & $5.13_{\pm1.64}$ \\
  & All-in-One & $37.29_{\pm5.81}$ & $28.56_{\pm6.62}$ & $53.11_{\pm4.28}$ & $22.74_{\pm10.09}$ & $24.88_{\pm7.72}$ & $29.51_{\pm8.89}$ & $44.84_{\pm14.42}$ & $25.32_{\pm16.76}$ & $24.47_{\pm11.38}$ & $21.37_{\pm4.07}$ & $4.42_{\pm2.77}$ \\
  & GPF & $41.55_{\pm9.18}$ & $30.57_{\pm9.77}$ & $57.96_{\pm5.07}$ & $37.79_{\pm6.15}$ & $23.72_{\pm10.19}$ & $39.47_{\pm8.46}$ & $58.46_{\pm15.43}$ & $20.90_{\pm4.09}$ & $27.58_{\pm17.99}$ & $23.48_{\pm8.69}$ & $6.36_{\pm1.65}$ \\
  & GPF-plus & $44.54_{\pm6.05}$ & $32.51_{\pm11.75}$ & $58.97_{\pm4.31}$ & $33.13_{\pm5.09}$ & $24.13_{\pm6.64}$ & $42.99_{\pm4.02}$ & $57.19_{\pm19.09}$ & $20.18_{\pm5.75}$ & $24.97_{\pm15.40}$ & $23.73_{\pm8.91}$ & $5.86_{\pm2.04}$ \\
  & SpikingGPF & $45.92_{\pm2.03}$ & $38.28_{\pm5.66}$ & $60.56_{\pm4.17}$ & $41.15_{\pm11.78}$ & $30.36_{\pm9.68}$ & $45.25_{\pm10.54}$ & $70.33_{\pm4.87}$ & $27.30_{\pm12.94}$ & $33.54_{\pm13.78}$ & $23.48_{\pm3.43}$ & $8.84_{\pm2.34}$ \\
\midrule
\multirow{6}{*}{\rotatebox{90}{\textbf{GraphMAE}}}
  & Fine-Tuning & $39.68_{\pm4.95}$ & $30.57_{\pm10.80}$ & $58.03_{\pm5.16}$ & $39.79_{\pm8.27}$ & $26.58_{\pm8.73}$ & $50.30_{\pm4.77}$ & $63.95_{\pm9.39}$ & $21.35_{\pm6.67}$ & $24.97_{\pm17.28}$ & $23.23_{\pm6.76}$ & $4.85_{\pm1.05}$ \\
  & All-in-One & $40.46_{\pm11.24}$ & $28.56_{\pm6.62}$ & $52.37_{\pm6.24}$ & $40.24_{\pm12.78}$ & $28.30_{\pm7.63}$ & $41.53_{\pm10.35}$ & $31.92_{\pm9.78}$ & $20.09_{\pm5.30}$ & $26.58_{\pm18.31}$ & $18.01_{\pm4.76}$ & $5.00_{\pm3.89}$ \\
  & GPF & $46.44_{\pm5.47}$ & $34.25_{\pm13.38}$ & $59.47_{\pm4.70}$ & $47.51_{\pm8.12}$ & $31.40_{\pm7.23}$ & $61.77_{\pm5.34}$ & $72.63_{\pm7.88}$ & $19.01_{\pm8.21}$ & $25.96_{\pm16.22}$ & $24.35_{\pm7.65}$ & $7.88_{\pm1.82}$ \\
  & GPF-plus & $45.46_{\pm3.75}$ & $33.90_{\pm13.05}$ & $58.79_{\pm3.70}$ & $47.38_{\pm6.91}$ & $31.33_{\pm7.95}$ & $62.39_{\pm5.75}$ & $74.09_{\pm8.02}$ & $19.37_{\pm8.51}$ & $25.34_{\pm18.08}$ & $23.85_{\pm7.83}$ & $7.54_{\pm1.51}$ \\
  & SpikingGPF & $49.30_{\pm5.10}$ & $40.35_{\pm8.54}$ & $61.71_{\pm5.09}$ & $47.19_{\pm7.94}$ & $35.49_{\pm9.29}$ & $64.53_{\pm6.01}$ & $75.91_{\pm7.30}$ & $20.54_{\pm9.24}$ & $25.96_{\pm15.61}$ & $25.22_{\pm7.22}$ & $10.26_{\pm1.79}$ \\
\midrule
\multirow{6}{*}{\rotatebox{90}{\textbf{EdgePred}}}
  & Fine-Tuning & $38.12_{\pm6.53}$ & $31.11_{\pm9.80}$ & $57.86_{\pm5.56}$ & $34.16_{\pm8.49}$ & $32.62_{\pm9.60}$ & $49.48_{\pm5.96}$ & $54.75_{\pm18.42}$ & $20.36_{\pm6.88}$ & $20.75_{\pm14.28}$ & $23.98_{\pm6.63}$ & $7.08_{\pm0.87}$ \\
  & All-in-One & $35.38_{\pm10.26}$ & $30.10_{\pm6.70}$ & $52.16_{\pm6.45}$ & $33.68_{\pm4.34}$ & $30.07_{\pm11.85}$ & $35.04_{\pm9.46}$ & $30.66_{\pm8.72}$ & $23.87_{\pm6.62}$ & $31.68_{\pm18.38}$ & $25.09_{\pm11.96}$ & $7.00_{\pm2.97}$ \\
  & GPF & $43.99_{\pm4.71}$ & $33.01_{\pm10.51}$ & $58.22_{\pm3.73}$ & $41.26_{\pm7.94}$ & $26.03_{\pm8.80}$ & $60.43_{\pm8.44}$ & $75.07_{\pm7.05}$ & $18.20_{\pm6.33}$ & $28.45_{\pm17.42}$ & $24.22_{\pm5.75}$ & $7.60_{\pm2.40}$ \\
  & GPF-plus & $43.23_{\pm2.12}$ & $33.15_{\pm11.32}$ & $59.29_{\pm5.62}$ & $40.42_{\pm4.99}$ & $27.61_{\pm9.50}$ & $60.25_{\pm6.71}$ & $70.00_{\pm8.37}$ & $19.01_{\pm5.71}$ & $25.47_{\pm17.58}$ & $23.35_{\pm8.36}$ & $6.82_{\pm1.91}$ \\
  & SpikingGPF & $51.81_{\pm5.21}$ & $38.13_{\pm8.38}$ & $62.60_{\pm6.03}$ & $45.10_{\pm7.96}$ & $34.64_{\pm9.94}$ & $63.32_{\pm6.54}$ & $74.01_{\pm6.17}$ & $24.14_{\pm8.42}$ & $29.32_{\pm14.62}$ & $24.47_{\pm4.67}$ & $10.62_{\pm2.07}$ \\
\midrule
\multicolumn{2}{c|}{GraphPrompt} & $50.20_{\pm9.16}$ & $32.29_{\pm11.38}$ & $56.29_{\pm4.90}$ & $45.85_{\pm9.41}$ & $33.45_{\pm11.03}$ & $55.60_{\pm5.21}$ & $58.58_{\pm11.35}$ & $18.20_{\pm5.64}$ & $28.32_{\pm14.70}$ & $25.84_{\pm9.48}$ & $10.18_{\pm2.92}$ \\
\midrule
\multicolumn{2}{c|}{GPPT} & $40.22_{\pm18.40}$ & $25.53_{\pm3.48}$ & $50.97_{\pm3.32}$ & $62.85_{\pm8.70}$ & $40.17_{\pm15.58}$ & $59.97_{\pm5.82}$ & $76.73_{\pm10.68}$ & $22.25_{\pm7.86}$ & $29.44_{\pm14.81}$ & $20.25_{\pm7.56}$ & OOM \\

\bottomrule
\end{tabular}
\end{adjustbox}
\end{table*}

\textbf{Implementation Details.}
 To ensure  the fair comparison, we adopt the widely used 3-layer GCN~\cite{kipf2016semi} as the  backbone with the hidden dimension of $256$ for all methods and the batch size is set to $128$. We use four pre-training methods (GraphCL~\cite{you2020graph}, SimGRACE~\cite{xia2022simgrace}, EdgePred~\cite{liu2023graphprompt} and GraphMAE~\cite{hou2022graphmae}) on the Flickr~\cite{zeng2019graphsaint} dataset. Then, we adjust the feature dimension of downstream data  to 100 and  then fine-tune the pre-trained models on the  downstream tasks. 
We leverage Adam~\cite{Kingma2014AdamAM} as the optimizer, with weight decay $4e-6$ and learning rate $1e-3$. The hyper-parameter $k$ of GPF-plus and our SpikingGPF is set to $10$. We tune the parameter $T$ from $\{1, 2, 4, 8\}$ and the firing threshold from $\{0.005, 0.05, 0.1, 0.2, 0.3\}$ based on the validation set.
All experiments are implemented using the ProG library~\cite{zi2024prog} which provides the implementations of numerous pre-training and prompting methods. All experiments are conducted on one single NVIDIA RTX 3090 GPU with 24GB memory. 

\subsection{Comparison Results}
We compare SpikingGPF against state-of-the-art methods as follows:  
{Supervised GNNs}: GCN~\cite{kipf2016semi} and GAT~\cite{velickovic2017graph} are trained directly on downstream labels in the supervised manner. 
{Graph tuning models}: fine-tuning method GPPT~\cite{sun2022gppt}, GraphPrompt~\cite{liu2023graphprompt}, All-in-one~\cite{sun2023all}, GPF~\cite{fang2023universal} and GPF-plus~\cite{fang2023universal}. 
Table \ref{tab:node_1shot} reports the results of all comparison methods. Here, we can observe that:
(1) Comparing with traditional supervised methods, most graph prompting approaches achieve better results, indicating that prompt learning can effectively exploit the latent knowledge of pre-trained model  to enhance the model’s adaptability on the downstream tasks. 
(2) Comparing with baseline models (GPF~\cite{fang2023universal} and GPF-plus~\cite{fang2023universal}), SpikingGPF gains consistent improvements, which  indicates the effectiveness of the proposed sparse feature prompting  by incorporating the sparse constraint via spiking neuron learning mechanism. 
(3) Comparing with other graph prompting methods, our SpikingGPF method achieves the best (or second-best) performance on all datasets under various pre-trained models,  validating its powerful competitiveness and effectiveness in graph prompt tuning problem.

\subsection{Model Analyses}
\subsubsection{Robustness Results}
\begin{figure*}[!ht]
    \centering
    \includegraphics[width=0.8\linewidth]{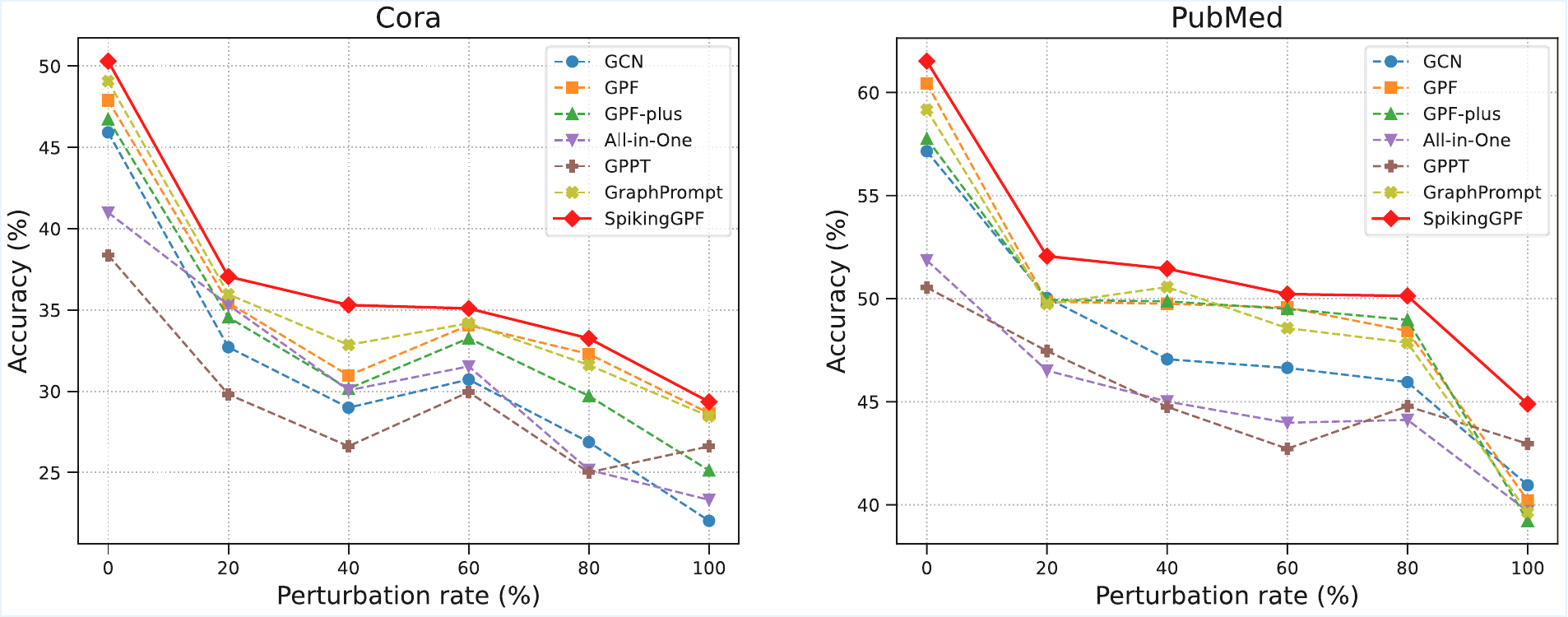}
    \caption{Comparison results of different methods under Random Attack on Cora and PubMed~\cite{sen2008collective} datasets with perturbation rate ranging from $20\%$
to $100\%$.}
    \label{fig:random_attack}
\end{figure*}

\begin{figure*}[!ht]
    \centering
    \includegraphics[width=0.8\linewidth]{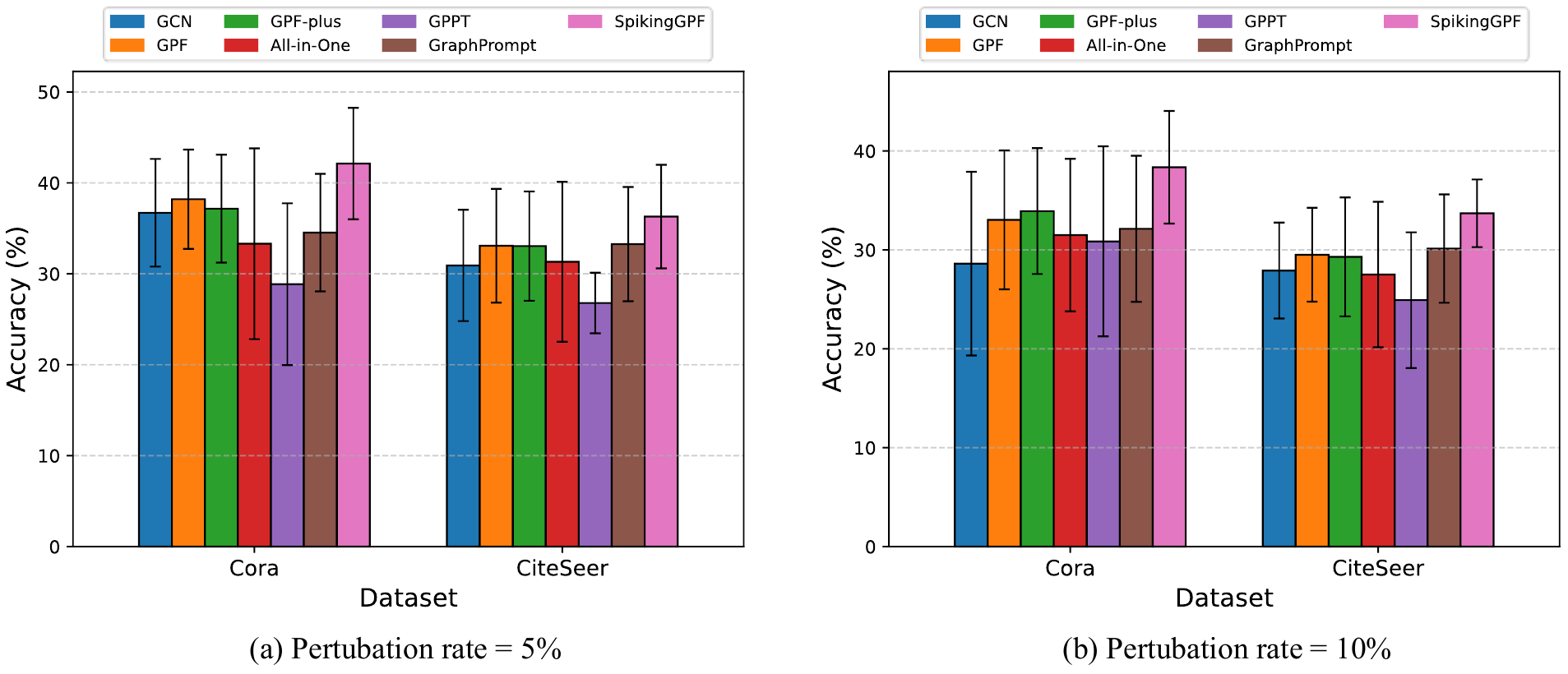}
    \caption{Comparison results of different methods under Metattack Attack on Cora and CiteSeer~\cite{sen2008collective} datasets when setting the perturbation rate to 5\% and 10\% respectively.} 
    \label{fig:metattack}
\end{figure*}

To evaluate the robustness of SpikingGPF, we conduct experiments on noisy graphs generated by Random Attack~\cite{li2020deeprobust} and Metattack~\cite{Metattack} respectively. Specifically, we use Random Attack with edge perturbations ranging from $20\%$ to $100\%$, and use Metattack to perform structural and feature attacks with perturbation rates of $5\%$ and $10\%$. Figure~\ref{fig:random_attack}-\ref{fig:metattack} show the results under different attack settings. Here, we can observe that (1) Comparing with GPFs, SpikingGPF achieves obviously better performance with different perturbation levels, demonstrating that our method enhances robustness against noisy data by incorporating the sparse constraints. (2) Comparing with other prompting methods, such as GPPT~\cite{sun2022gppt}, All-in-One~\cite{sun2023all} and GraphPrompt~\cite{liu2023graphprompt}, SpikingGPF consistently obtains better results,  demonstrating the robustness of our spiking graph prompt learning. 

\subsubsection{Ablation Study}
\begin{figure}[!htp]
    \centering
    \includegraphics[width=0.95\linewidth]{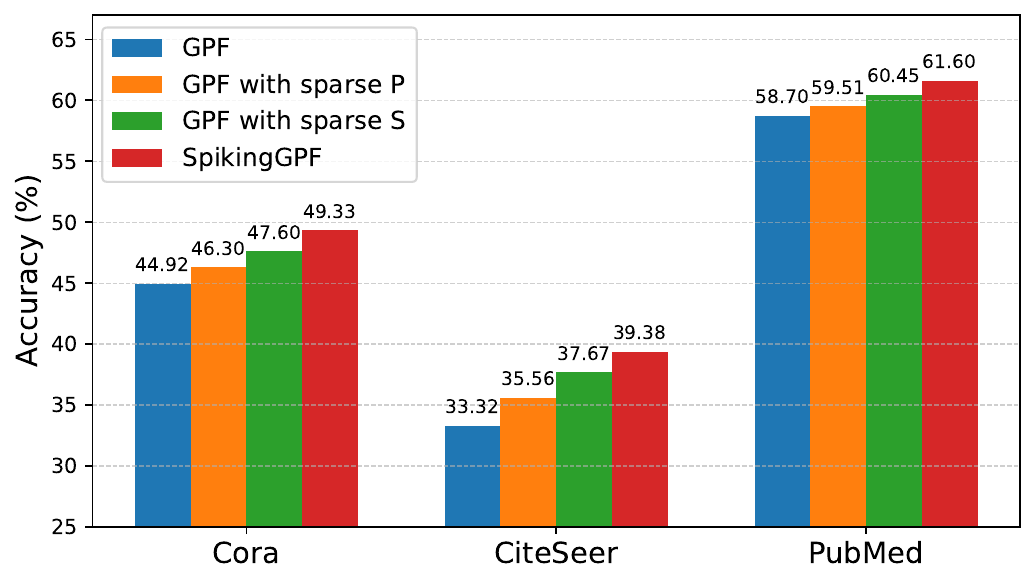}
    \caption{Ablation analysis of the proposed SpikingGPF on three datasets. The two variants obtained by augmenting the baseline with sparse \textbf{S} only and sparse \textbf{P} only consistently achieve stable improvements across all datasets. SpikingGPF with two sparse mechanisms achieves the best performance.}
    \label{fig:Ablation}
\end{figure}

Our SpikingGPF contains two main components, \textit{i.e.}, \textbf{S}-learning module and \textbf{P}-learning module. To evaluate the contribution of each component, 
we conduct an ablation study by removing these components. The ablation study results are summarized in Fig.~\ref{fig:Ablation}. We can observe that  (1) SpikingGPF with two sparse mechanisms achieves the best performance on all datasets with a substantial margin. 
(2) The two variants obtained by augmenting the baseline with sparse \textbf{S} only and sparse \textbf{P} only consistently achieve stable improvements across all datasets, demonstrating the effectiveness of the two components.

\subsubsection{Parameter Analysis}
\begin{figure*}[!ht]
    \centering
    \subfigure[Parameter analysis of learned $\mathbf{S}$ ]{
        \includegraphics[width=0.48\linewidth]{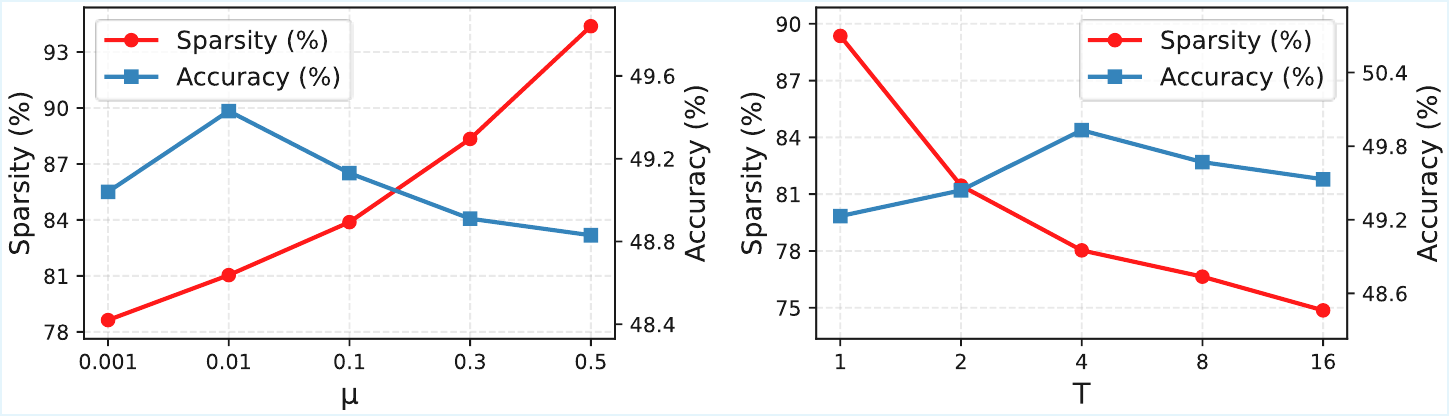}
    }
    \subfigure[Parameter analysis of learned $\mathbf{P}$ ]{
        \includegraphics[width=0.48\linewidth]{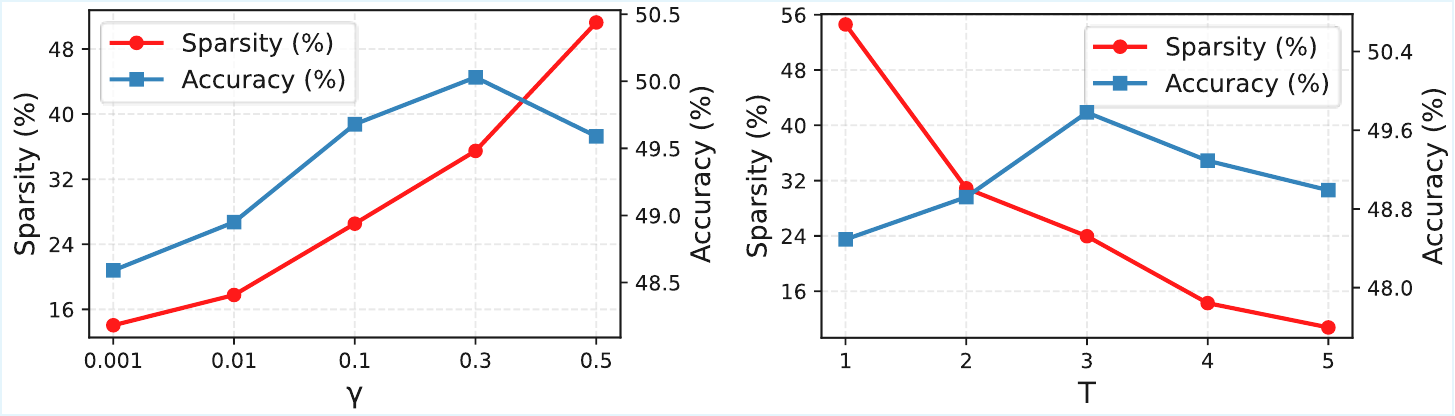}
    }
   \caption{Sparsity of the learned $\{\mathbf{S}$, $\mathbf{P}\}$ and performance of SpikingGPF with different threshold $\{\mu, \gamma\}$ and $T$ on Cora~\cite{sen2008collective} dataset. We can observe that larger $\mu$ and $\gamma$ encourage sparser $\mathbf{S}$ and $\mathbf{P}$ while large $T$ leads to denser $\mathbf{S}$ and $\mathbf{P}$.}
    \label{fig:Parameter_analysis}
\end{figure*}

As mentioned in Section~\ref{Method}, the sparsity of \textbf{S} and \textbf{P} is controlled by firing threshold ($\mu$, $\gamma$) and parameter $T$. Therefore, we analyze how these parameters affect the sparsity of \textbf{S} and \textbf{P} and model's performance. As shown in Fig.~\ref{fig:Parameter_analysis},  
we can observe that:
(1) larger $\mu$ and $\gamma$ encourages  sparser \textbf{S} and \textbf{P} while smaller $T$  leads to sparser \textbf{S} and \textbf{P}. (2) Excessive sparsity may lose some useful information and thus results in worse results. Therefore, we can take a trade-off for the best performance. 
%
%

\subsubsection{Performance with Different Shots}
\begin{figure}[!htbp]
    \centering
    \includegraphics[width=\linewidth]{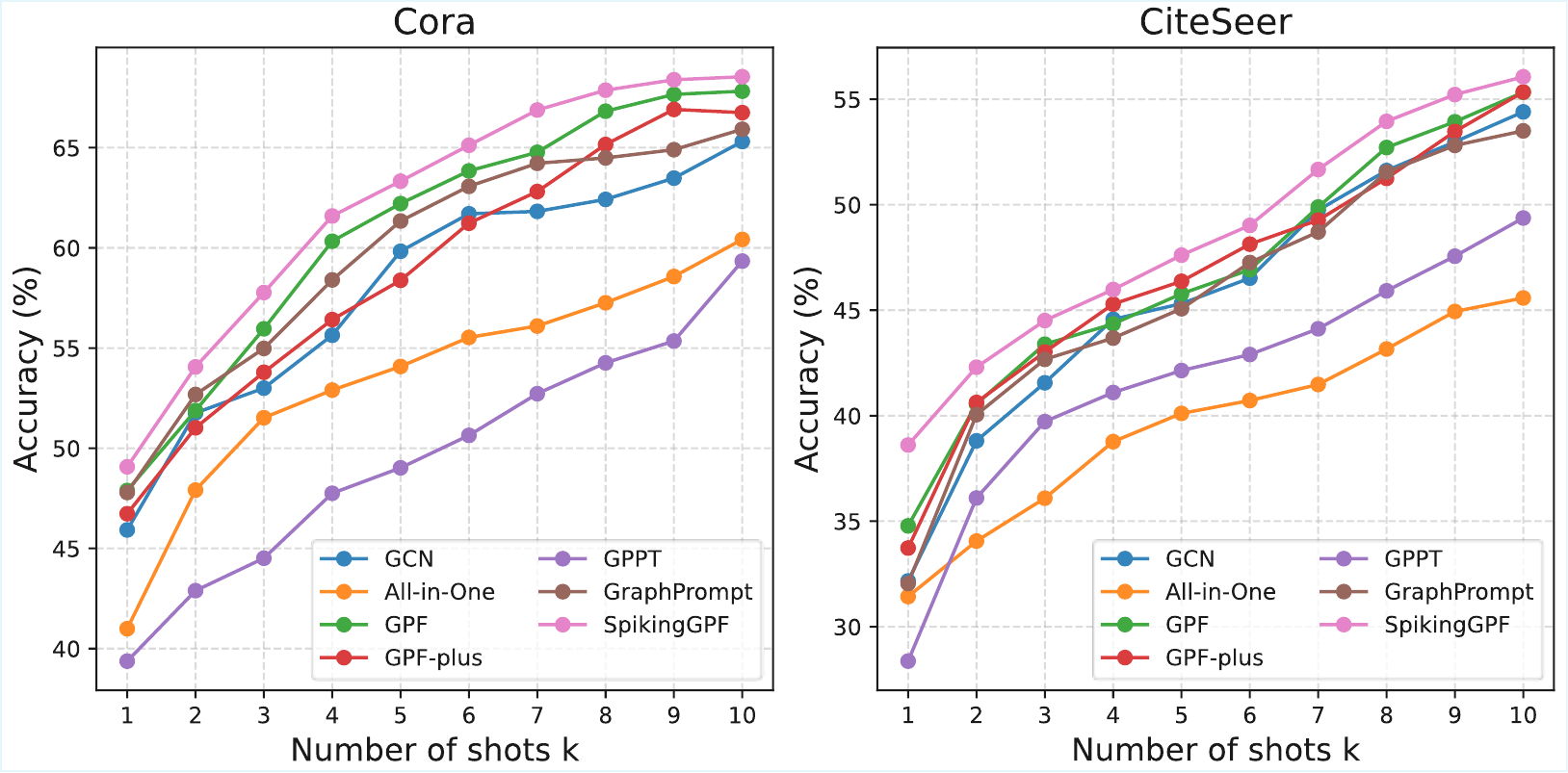}
    \caption{Impact of shots on few-shot node classification. We can find that, the performance of all methods improves as the number of shots increases, and our proposed SpikingGPF consistently outperforms compared baselines across different shots.}
    \label{fig:shot_number}
\end{figure}

To further investigate the impact of shots, we vary the number of shots from $1$ to $10$ and show the results 
in Fig.~\ref{fig:shot_number}. Overall, the performance of all methods improves as the number of shots increases. Moreover, our proposed SpikingGPF consistently outperforms compared methods across different shots, which further demonstrates the advantage of the proposed SpikingGPF method. 

\section{Conclusion}
This paper presents a novel graph prompting method, termed Spiking Graph Prompt Feature (SpikingGPF) by integrating sparse constraints into GPF architecture. 
By exploiting the mechanism of spiking neuron, SpikingGPF learns a sparse coefficient \textbf{S} which selectively combines a set of basis prompts to yield a more compact and efficient prompt representation. 
Moreover, SpikingGPF exploit signed IF neuron mechanism to learn sparse $\textbf{p}_i$ for each graph node, yielding a more
compact and lightweight prompting design while obviously improving robustness against various graph node noise. 
Extensive experiments on various benchmark datasets demonstrate the effectiveness and robustness of the proposed SpikingGPF.

\bibliographystyle{IEEEtran}
\bibliography{reference}

\end{document}